\pdfoutput=1

\documentclass[11pt]{article}

\usepackage{ACL2023}

\usepackage{times}
\usepackage{latexsym}

\usepackage[T1]{fontenc}

\usepackage[utf8]{inputenc}

\usepackage{microtype}

\usepackage{inconsolata}

\usepackage{amsmath}
\usepackage{graphicx}

\DeclareMathOperator{\funcidf}{idf}
\DeclareMathOperator{\functf}{tf}

\title{Future Sight: Dynamic Story Generation with Large Pretrained Language Models}

\author{Brian D. Zimmerman, Gaurav Sahu, Olga Vechtomova \\
  University of Waterloo\\
  Waterloo, ON, Canada \\
  \texttt{\{bdzimmer,gsahu,ovechtom\}@uwaterloo.ca}
  }

\begin{document}
\maketitle
\begin{abstract}
Recent advances in deep learning research, such as transformers, have bolstered the ability for automated agents to generate creative texts similar to those that a human would write. By default, transformer decoders can only generate new text with respect to previously generated text. The output distribution of candidate tokens at any position is conditioned on previously selected tokens using a self-attention mechanism to emulate the property of autoregression. This is inherently limiting for tasks such as controllable story generation where it may be necessary to condition on future plot events when writing a story. In this work, we propose \textit{Future Sight}, a method for finetuning a pretrained generative transformer on the task of future conditioning. Transformer decoders are typically pretrained on the task of completing a context, one token at a time, by means of self-attention.  \textit{Future Sight} additionally enables a decoder to attend to an encoded future plot event. This motivates the decoder to expand on the context in a way that logically concludes with the provided future. During inference, the future plot event can be written by a human author to steer the narrative being generated in a certain direction. We evaluate the efficacy of our approach on a story generation task with human evaluators.

\end{abstract}

\begin{table}[h]
  \centering
  \begin{tabular}{p{1.5cm} p{5.5cm}}
  \hline
  \textbf{Context} & 
    \small
    ``No! I refuse to let you do this!''
    \vspace{0.2cm}
    
    I turned to look at my wife's face, which was covered in the glistening drops of sorrow that were tears, her eyes red and voice hoarse from pleading and crying,
    \vspace{0.2cm}

    ``Please don't go, there has to be another way to fix it!'' 
    \vspace{0.2cm}

    ``Look around us!'' I yelled back, gesturing wildly, 
    \vspace{0.2cm}
    
    ``Look what I did!
    \normalsize
    \\
  \hline
  \textbf{Future} &
    \small
    The swamp creatures were relentless in their siege.
    \normalsize
    \\
  \hline
  \textbf{Future Distance} &
    \small
    3 sentences
    \normalsize
    \\
  \hline
  \textbf{Prediction} &
    \small
    This is our town! We've been \textbf{in the mud} since the first \textbf{raiders!} We haven't been able to \textbf{stop the invaders} and their \textbf{hordes}.
    \normalsize
    \\
  \hline
  \end{tabular}
  \caption{\label{tab:swampstory} Example story from \textit{Future Sight} conditioned on an impending swamp creature attack. Bolded words denote the perceived effects of the conditioning.}
\end{table}

\section{Introduction}
Linguistic expression is but one modality of human creativity, and therefore difficult to emulate. The story is one example of linguistic expression that is greater than the sum of its parts. More than just a collection of words, the story must be carefully crafted. It’s essential for the storyteller to simultaneously manage many personae, events, and emotions which are each drawn from experience. The storyteller must maintain a minimum degree of coherence and consistency to keep their narrative immersive.

With the introduction of transformers and self-attention~\citep{vaswani2017attention}, the generation of near-human quality longform texts became possible. Self-attention is a mechanism for learning an alignment between a token in an ordered sequence with every other token in the sequence. Solely relying on self-attention, transformers set several state-of-the-art benchmarks in generative language modeling \citep{radford2018improving, radford2019language, brown2020language} and representation learning \citep{devlin2018bert}.

In generative language modeling tasks, transformers are notoriously difficult to control. This is due to the fact that they emulate the autoregressive property of conditional sequence decoding through the use of attention masking. The attention mask disallows the model from attending to future tokens in order to prevent future knowledge from influencing the selection of a token at the present timestep.

There are zero-shot techniques, such as prompt tuning, which affix instructions to the beginning of the context in an effort to influence generated text \cite{radford2019language}. This is problematic because these instructions effectively become part of the context and cannot easily be changed, something that may be necessary quite frequently in a story with many twists and turns. Alternatively, a future event could be fed to the decoder with the context, separated only by masked tokens. This imposes a constraint on the model by limiting the number of tokens between the context and the future event.

We modify a pretrained GPT-2 implementation from \citet{wolf2019huggingface} to optionally attend to future events without the need for directly modifying the decoder context at any point. At inference time, the encoded future can be changed at any time to shift the direction of the story being generated. To this end, \textit{Future Sight} enables true dynamism in most creative text generation tasks. We anticipate human authors will be able to collaborate with our proposed system to write stories by providing key events in a plot and allowing the model to generate intermediate details via temporal interpolation. By providing key plot events in the future narrative, a human author can guide the system towards these events without the need to re-write the past narrative. In this way, the generated narrative can be steered by the human author as the story develops instead of being planned in advance. 

Our main contributions are as follows:

\begin{itemize}
    \item We propose the task of \textit{future conditioning} – the process of coalescing a context sequence and a future representation by motivating a generative language model to predict intermediate sequences that are consistent with both context and future.
    \item We release \textit{Future Sight}, our implementation of future conditioning on a story generation dataset using transformers and publicly available pretrained weights.
    \item We propose mean-idf as a heuristic for selecting future plot events with high information capacity in support of training \textit{Future Sight}.
\end{itemize}

\section{Background}
There are many prior works on automated story generation and controlling transformer decoders, but no prior approaches that combine the two without explicitly pretraining on a story generation task. With \textit{Future Sight}, we choose to utilize pretrained transformers because they are easy to finetune and have the ability to generate coherent long texts out of the box.

\subsection{Automated Story Generation}
Automated story generation is the task of using either a set of heuristics~\cite{ li2013story,meehan1977tale} or deep neural networks~\cite{fan2018hierarchical, rashkin2020plotmachines, yao2019plan} to exhibit attributes of \textit{narrative intelligence} in story writing. \citet{riedl2006linear} describe narrative intelligence as the ability of a storytelling agent, human or machine, to convey an experience as a story. Narrative intelligence encapsulates qualities of stories such as thematic and syntactic consistency, two attributes which come naturally to humans yet remain nontrivial for automated storytelling agents to emulate.

Heuristic approaches were the primary means to developing an automated story generation agent before deep neural networks were mature enough to generate longform texts. One of the earliest heuristic approaches to automated story generation mapped user input to sequences of rules from which a story could be constructed~\citep{meehan1977tale}. More recently, \citet{li2013story} formed tuples of events and used crowdsourcing to create graphs describing the precedence and mutual exclusion relationships between them. Heuristic methods require extensive use of humans-in-the-loop to develop and are only capable of producing a finite set of stories. \textit{Future Sight} is built with transformers to enable the generation of many stories from one datapoint without any costly human intervention.

\citet{fan2018hierarchical} introduced the idea of prompt guided story generation through a hierarchical model. The \textit{cold fusion} approach uses two convolutional neural networks to focus on different abstractions of the story as it relates to a prompt. One network focuses on the interdependencies of the produced story, ensuring that it remains coherent and sensible throughout. The second convolutional neural network is trained at a later point, to accommodate any higher-level prompting information. While \citet{fan2018hierarchical} demonstrated a prompt-based conditioning effect with their model, prompt-based conditioning doesn't allow for any fine control over the generative process. \textit{Future Sight} integrates specific plot events as complete sentences into a story as part of an interactive process.

\citet{yao2019plan} utilize a two-step alternating process of planning and writing. A one-word plot event conditions the next sentence in a story. From the generated sentence, a subsequent plot event can be produced, continuing until story completion. Although this approach was one of the first to introduce a notion of dynamism in the form of planning, it faced limitations of its own. Plot events were produced by a recurrent planning mechanism, which places limitations on the user at inference time. Additionally, one word plot events contain very little critical information. There is evidence to suggest that events are higher order components of stories that require more robust representations \cite{li2013story,martin2018event}. \textit{Future Sight} addresses these limitations by enabling the generative model to attend to a plot event as a complete sentence with the expectation that will appear somewhere in the story later on.

\citet{rashkin2020plotmachines} use transformers to generate complete paragraphs of text according to a provided story outline. By augmenting a transformer decoder with a memory mechanism, \textit{Plotmachines} can track which parts of a plot have yet to be addressed when generating subsequent paragraphs. This approach trains a transformer decoder from scratch, a process that is expensive and results in a model that may have trouble generalizing to tasks in other domains. \textit{Future Sight} requires no expensive pretraining procedure and doesn't rely on augmenting the context with complex flags.

\subsection{Controlling a Transformer}
A caveat of using pretrained transformers is that some downstream tasks require signalling that is not accounted for during model pretraining. Moreover, some tasks require explicit instructions that cannot be inferred by a model through zero-shot task transfer as demonstrated by models such as GPT-2~\cite{radford2019language} and GPT-3~\cite{brown2020language}. For tasks such as story generation, the prompt must be modified each time the user wishes to steer the narrative in a specific direction, requiring recalculating attention at each position. \textit{Future Sight} addresses this limitation by decoupling context and conditioning information, instead opting to incorporate conditioning information as a new temporal dimension in the self-attention module of the decoder. This allows the user to steer the narrative without changing the prompt (context), which is a desirable feature for collaborative human-machine story writing.

Attempts to guide transformers by redefining the pretraining procedure to accommodate additional domain-specific flags saw some success. \citealp{keskar2019ctrl} prepend model contexts with a one word token selected from a finite list of domain labels called \textit{control codes}. Control codes in CTRL correspond to the source of each dataset used during pretraining which allows the model to learn an alignment between the language in a source sequence and the domain from which the source sequence was drawn. The use of varying control codes led to a quantifiable and predictable difference in generation when passing the same prompt during inference. Unfortunately, CTRL requires an expensive pretraining procedure, something we sought to avoid with \textit{Future Sight}.

\citet{krause2020gedi} avoid the costly pretraining procedure from CTRL without obfuscating the weights of the generative language model by using generative discriminators (GeDi). This process requires two GPT-2 instances, one serving as a generative language model while the other is finetuned as a class-conditional language model (CCLM). The CCLM is provided a conditioning flag and a context in order to produce a perturbation over the output distribution of the generative language model. Techniques which perturb output logits directly, have been shown to degenerate the fluency of generated texts. Our model avoids this altogether by injecting conditioning information directly into the self-attention mechanism of the transformer decoder.

\citet{li2020optimus} learn a smooth latent space from which to sample vectors for transformer decoder conditioning. By connecting BERT and GPT-2 as a \textit{variational autoencoder}, BERT learns a continuous distribution of encoded sentences rather than just one embedding for a source sentence. A pretrained GPT-2 is modified to ingest any embedding sampled from this continuous space and dynamically condition on it without the need to deviate from the structure of the pretraining data. Sampled vectors can be attended to as position zero, a process referred to as \textit{memory injection}. In \textit{Future Sight}, we modify this approach and suit it to our \textit{future conditioning} task.

\section{Methodology}
The primary difference between our approach and prior techniques in automated story generation is the ability for users, such as authors, to directly interact with the generative process by providing new future plot events at any time. We first introduce the task of future conditioning.

\begin{figure}
    \centering
    \includegraphics[scale=0.4]{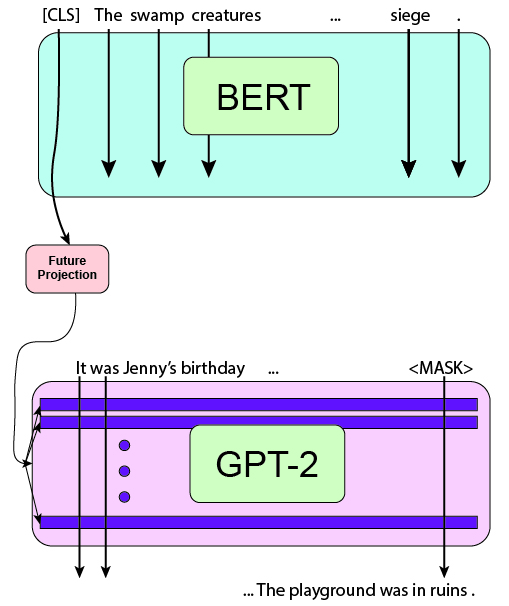}
    \caption{An end-to-end depiction of \textit{Future Sight}. A future event involving swamp creatures is injected into the self-attention module of GPT-2 as it generates the events of Jenny's birthday party.}
\end{figure}

\subsection{Future Conditioning Task}
We define the task of \textit{future conditioning} as guiding a generative language model to interpolate between a context $c$ and a future event $f$. Interpolation, in this sense, refers to moving between two events temporally rather than interpolating between two points in a continuous space. With respect to story generation, future conditioning can be thought of as smoothly blending together a context paragraph and a future plot event, each encoded as complete sentences. From the end of a context paragraph $c$, the model should prioritize selecting tokens that both expand on $c$ and logically conclude with a sentence $f$. During inference, the user can specify plot events to continue the story, with or without regard to the text that the model has previously generated.

From the set of all stories $S$, we begin by running a sentence splitting function on each $s\in{S}$. We use the first $\ell_{c}$ sentences in each resultant list as the contexts for our data points. Each subsequent $\ell_{f}$ sentences represent a collection of \textit{future candidates} for a datapoint, from which we must nominate a future.

\subsubsection{Selecting a future}
It is important to select a future candidate with a high information capacity in order to maximize the influence of the future on the generative process. We use the mean-idf heuristic to detect sentences with a high ratio of unique and informative words to select as a future. The future with the highest mean-idf score is nominated to represent the datapoint as a future event. Along with selecting a future, we record the distance, in sentences, that the selected future candidate is from the end of the context. This \textit{future distance} will be jointly encoded with the future event to provide the model with some notion of how soon the selected future candidate is to occur with respect to higher order temporal events.

\small

\begin{equation}
    \centering
    \funcidf(t,D) = \log{\frac{\sum\limits_{d\in{D}}1}{1 + \sum\limits_{d\in{D}}\begin{cases}
    1,& \text{if } \functf(t,d) > 0\\
    0,& \text{if } \text{otherwise}
    \end{cases}}}
    \label{eqn:idf} 
\end{equation}
\normalsize

\subsubsection{The Mean-IDF heuristic}
\citet{jones1972statistical} proposed inverse document frequency (IDF) as a measure of a term's informativeness in a corpus of documents. The IDF for a term $t$ is computed as the logarithmic ratio of the total number of documents in a corpus $N$ to the number of documents containing at least one instance of $t$ (see Equation~\ref{eqn:idf}). 

We devise a heuristic for selecting informative futures by using averages over the IDF score of each term $t$ in a future candidate $f$. Our corpus $D$ is the set of all sentences extracted from each $s\in{S}$. We deliberately include sentences from each story context in this corpus. Intuitively, it is less imperative for a future embedding to contain information that is already contained in the story context so we should penalize such futures. Each future candidate is first tokenized into a list of words with stopwords excluded. We average the IDF values of all words in the resultant list for each future candidate, nominating the highest scorer for our future event. Future candidates with rarer words, and thus higher scores, we presume to contain more information to pass to the decoder.

\subsubsection{Story Reconstruction}
\textit{Future Sight} is tasked with reconstructing the story $s$ from the context $s_{[1:\ell_{c}]}$ and the encoded future $f$. We use a cross-entropy objective over the generated tokens to finetune  \textit{Future Sight} on the future conditioning task.

\subsection{Model Architecture}
Though our model requires both an encoder and a decoder, it's not an encoder-decoder model in the traditional sense as transformer decoders are capable of generating on their own due to their self-attention mechanism.

\subsubsection{Encoder}
An encoder is necessary for creating a high level representation of the future event. The representation created by the encoder should contain enough information to influence the decoder while generating sentences up to the time at which the future should occur.

Future events in our story generation task are represented as complete sentences. We utilize a popular transformer encoder, BERT, with publicly available pretrained weights to encode our future event. We concatenate the future distance and the tokenized future with a separator token \textbf{SEP} before feeding it through BERT. For our future representation, we use the output from BERT’s position 0. The output at this position serves as the classification head during the BERT pretraining procedure and naturally provides a strong representation of the input that can be utilized in downstream tasks.

\subsubsection{Future Injection}
Because Future Sight is designed to work with pretrained transformers, we refrain from deviating from the decoding procedure used by transformer and GPT-2. We instead opt for techniques to enable the decoder to optionally attend to the encoded future plot event at any time.

To integrate the encoded future plot event into the decoder, we referred to the embedding injection techniques from prior works. \citealp{li2020optimus} conditioned a GPT-2 instance on embeddings sampled from a continuous latent space as part of a variational autoencoder. These embeddings were injected into GPT-2 in two ways: memory injection and embedding injection.

\textit{Embedding injection} refers to concatenating conditioning information directly to the embedding of each input token in the decoder. Though we added support for embedding injections to our model, we didn't observe a conditioning effect while using it. Due to this, we focus on memory injection for the duration of this work.

\textit{Memory injection} concatenates the conditioning information to the self-attention module at each position. In this way, future information can be optionally referenced in computing the attention at each layer. As each decoder layer has its own self-attention weights focusing on different abstractions of the context features, we provide each layer with a different learnable projection of the encoded future as well. 

\subsubsection{Decoder}
For our pretrained decoder, we opt to use a Pytorch implementation of GPT-2 which was made available by HuggingFace. To accommodate the memory injection, we augment the GPT-2 implementation with an additional attention position in the layer module.

For the story generation task, we chose to encode each future with BERT base, an implementation with a 768 dimensional hidden space. This matches the dimensionality of the hidden space of the GPT-2 implementation we utilize in \textit{Future Sight}, but we still need to make one additional modification. We increase the dimensionality of the future representation to $12 * 768$ with a nonlinear transformation in order to create a distinct future vector for each layer in the decoder. By default, Huggingface GPT-2 has 12 layers and 117 million total parameters.

\section{Experiments}
\subsection{Dataset}
For our experiments, we use the WritingPrompts dataset~\cite{fan2018hierarchical}, which consists of 303,358 stories and prompts from r/WritingPrompts, a Reddit community oriented around story authorship and contribution. The goal of this community is to have users propose prompts for other users to respond to with a related story. We discard the prompts as our task is to condition GPT-2 on an event drawn from a source story itself rather than a general prompt. We additionally ignore stories that are less than 9 sentences long, resulting in 291,575 data points.

\subsection{Training Procedure}
We train our model on one Nvidia RTX 2080 for 5 epochs. Due to hardware limitations, we use a batch size of 1 and perform an optimization step every 16 batches for an effective batch size of 16. The entire procedure takes approximately 21 hours.

\subsection{Evaluation}
Evaluating our model proved tricky as there is a lack of metrics in prior works for evaluating creative tasks such as \textit{Future Sight}. Metrics which are typically used to evaluate generative language models, such as BLEU~\cite{papineni2002bleu}, were designed with translation tasks in mind. Translation emphasizes accuracy over diversity in generation which is counterintuitive for story generation. Additionally, there were no reliable metrics we could find in prior work with which to evaluate the conditioning effect of our model.

We developed a classification task to compare \textit{Future Sight} with a standard pretrained ``vanilla'' GPT-2 model of the same parameter specifications as our decoder. We construct a dataset consisting of ground truth contexts and futures along with predictions from both \textit{Future Sight} and the vanilla GPT-2 model for our task.

\begin{table}[h]
  \centering
  \begin{tabular}{p{1.5cm} p{5.5cm}}
  \hline
  \textbf{Context} & 
    \small
    ``Welcome back, Mr.~Jones''.  
    \vspace{0.2cm}
    
    I blinked at the bright light until my eyes were able to focus on the overly-cheerful blonde standing next to me. I ran my tongue over my lips to try getting some moisture onto them.
    \vspace{0.2cm}

    ``Where am I?'' I croaked.
    \normalsize
    \\
  \hline
  \textbf{Future} &
    \small
    ``I'm nurse Patkins, and I'm going to go over some stuff with you.''
    \normalsize
    \\
  \hline
  \textbf{Future Distance} &
    \small
    3 sentences
    \normalsize
    \\
  \hline
  \textbf{Prediction} &
    \small
    He looked around me as if it were trying to figure out where he was. I tried to focus on \textbf{the nurse}, but the sound of the door opening and closing stopped me.
    \vspace{0.2cm}

    ``Where am I?'' I croaked. ``I'm in recovery from a cardiac arrest, Doctor?''
    \normalsize
    \\
  \hline
  \end{tabular}
  \caption{\label{tab:truestory} Example story from \textit{Future Sight} where the presence of a nurse could be inferred from the ground truth future.}
\end{table}

\subsubsection{Human Evaluations}
We asked human evaluators to classify the predictions generated by each model. Provided with a context, a ground truth future, and predicted intermediate text, the evaluators were tasked to choose from three classes. The first two classes were continuations of the context as predicted by \textit{Future Sight}. The first \textit{Future Sight} class represented predictions that were conditioned on the ground truth future extracted from the same story containing the context, such as the story in \autoref{tab:truestory}. The second \textit{Future Sight} class represented predictions that were conditioned on the fixed future ``The swamp creatures were relentless in their siege.'', as in  \autoref{tab:swampstory}. This fixed future was chosen empirically as we observed the profound effect it had on conditioning that was easily identifiable. The final class represented examples which contained a predicted continuation of the context as generated by vanilla GPT-2. As vanilla GPT-2 cannot accommodate future information, the future is not used at inference time for examples in this category.

\begin{table}[]
    \small
    \centering
    \def\arraystretch{1.25}
    \begin{tabular}{|p{0.75cm}||p{1cm}|p{1cm}|p{1cm}|p{1cm}|}
        \hline
        \multicolumn{5}{|c|}{Class 1 --- \textit{Future Sight} (True Future)} \\
        \hline
        \textbf{Agg.}& \textbf{Acc.} & \textbf{Pr.} &\textbf{Re.} &\textbf{F1}\\
        \hline
        $\mu$       &0.4386 &0.3767 &0.4393 &0.3905\\    
        $\sigma$    &0.1793 &0.0725 &0.1792 &0.1001\\
        \hline        
        \multicolumn{5}{|c|}{Class 2 --- \textit{Future Sight} (Swampy Future)} \\
        \hline
        \textbf{Agg.}& \textbf{Acc.} & \textbf{Pr.} &\textbf{Re.} &\textbf{F1}\\
        \hline
        $\mu$       &0.5601 &0.9069 &0.5572 &0.6761\\    
        $\sigma$    &0.1184 &0.0784 &0.1188 &0.0717\\
        \hline        
        \multicolumn{5}{|c|}{Class 3 --- Vanilla GPT-2} \\
        \hline
        \textbf{Agg.}& \textbf{Acc.} & \textbf{Pr.} &\textbf{Re.} &\textbf{F1}\\
        \hline
        $\mu$       &0.6992 &0.6085 &0.6992 &0.6400\\    
        $\sigma$    &0.1089 &0.0687 &0.1090 &0.0369\\
        \hline        
        \multicolumn{5}{|c|}{Total --- Class Aggregated (Macro)} \\
        \hline
        \textbf{Agg.}& \textbf{Acc.} & \textbf{Pr.} &\textbf{Re.} &\textbf{F1}\\
        \hline
        $\mu$       &0.6066 &0.6312 &0.5653 &0.5951\\    
        $\sigma$    &0.0445 &0.0268 &0.0619 &0.0447\\
        \hline
    \end{tabular}
    \caption{\label{tab:humanresults} Aggregated results of the classification task given to six human evaluators. \textbf{Class 1} refers to generated stories conditioned on ground truth futures. \textbf{Class 2} refers to stories conditioned on the fixed future \textit{``The swamp creatures were relentless in their siege.''} \textbf{Class 3} refers to stories generated by a pretrained GPT-2 with no additional finetuning.}
\end{table}

\section{Results}

We report accuracy, precision, recall, and f1 score for each participant across each class in \autoref{tab:humanresults}. Of all of the computed metrics, perhaps most notable was the high precision among all participants for stories conditioned on the swamp future. This indicates that participants were very confident in their decision to label a story with class 2, seldom incorrectly assigning the label. Recall was the lowest among predictions conditioned on true futures. As class 2 had high reported precision, it's reasonable to assume that many instances of stories conditioned on true futures were incorrectly labeled with class 3. Class 3, in essence, represents a ``no conditioning'' class. Class 1 data points conditioned on weak and uninformative futures could easily be falsely attributed to vanilla GPT-2.

Feedback from the human evaluators was generally consistent. Participants each noted that it was difficult to discern Class 1 from Class 3 in many cases. Participants noted that this often led to guessing at random between these two classes. Differentiating between classes 1 and 3 was particularly difficult for participant 2, who labeled 76 out of 100 examples with Class 3. Despite this, there were some cases within the 100 examples that the evaluators found to be obviously conditioned on the true future.

\subsection{Discussion}

Though the results of the above classification tasks suggest that \textit{Future Sight} can successfully integrate a future event into a story, there are two key areas in which we believe it falls short. The first area is the lack of a good story dataset. We'll discuss the dataset we used for our experiments and how the results led us to this conclusion. The second area of improvement is the lack of a reliable metric to evaluate future conditioning.

\begin{table}
  \centering
  \begin{tabular}{p{1.5cm} p{5.5cm}}
  \hline
  \textbf{Context} & 
    \small
    It happened suddenly and at first I thought I had died. Finally I could sense something beyond the chemical sentences to yelled at me to deliver. I felt something. I felt. I was ripped away from the natural flow into something I never wanted or ever even thought to seek. \\
    \normalsize
    \\
  \hline
  \textbf{Future} &
    \small
    I found myself, I existed.
    \normalsize
    \\
  \hline
  \textbf{Future Distance} &
    \small
    3 sentences
    \normalsize
    \\
  \hline
  \textbf{Prediction} &
    \small
    I could feel it. And I was alone. I was in my room and I knew.
    \normalsize
    \\
  \hline
  \end{tabular}
\caption{\label{tab:ambiguousstory} Example of a story conditioned on an ambiguous future.}
\end{table}

\subsection{Dataset Problems}
Throughout the course of the development of \textit{Future Sight}, we've identified several issues with the WritingPrompts dataset which we believe have negatively impacted the results overall.

As a collection of uncurated stories from an online forum, there is extensive markup, overuse of punctuation, incomplete dialogue, and, bizarrely, text that has been modified to look unsettling by being obfuscated with unicode diacritics. Though some of these outliers are simple enough to catch and discard, some of them are nontrivial to deal with. For example, a future event containing long spans of punctuation and little substance will only stifle \textit{Future Sight} as it attempts to extract meaningful information from it.

Additionally, some of the stories in the WritingPrompts dataset are very ambiguous or philosophical such as the story in \autoref{tab:ambiguousstory}. For example, there are many stories written as first-person narratives containing the ``stream of consciousness'' literary device. Though occasionally interesting, narratives styled in this way often lack any expository details regarding an overall plot for the model to focus on. As \textit{Future Sight} attempts to condition on a future extracted from such a story, it becomes unclear to readers whether or not the conditioning effect is present at all. Additionally, there are not many stories where a future sentence contains a concise description of an important plot event or a character that cannot be inferred from the context alone, such as the example given in \autoref{tab:truestory}.

\subsection{Lack of Automated Metrics}
Creative tasks in natural language generation are notoriously difficult to evaluate with automated metrics. Based on the results of our experiment, we conclude that there is a lack of reliable automated metrics for evaluating our proposed task. Though the results of the human classification task may indicate the presence of discernible conditioning in stories generated by \textit{Future Sight}, measuring the strength of the conditioning itself is nontrivial. Additionally, automating the evaluation of fluency and premise continuity has always been difficult for story generation researchers. Though human evaluators can provide an assurance that what is being written is seemingly legitimate, humans can be unreliable. Moreover, as stories are often deeply intertwined with culture, results could vary by the region surveyed.

\section{Conclusion}
Here we summarize some directions for future work as well as reiterate a summary our contributions.

\subsection{Future Work}
Throughout the course of this project, we've identified several areas for future work that can expand on the future conditioning task.

\noindent \textbf{Automated Metrics}
As previously stated, development of an automated metric for evaluating the effects of \textit{future conditioning} would be nontrivial. We have considered looking at works within information retrieval and document ranking to evaluate similarities between predicted intermediate sentences and the future.

\noindent \textbf{Strengthening Future Importance}
Depending on the context and the future, the effects of \textit{Future Sight} conditioning may not be obvious. We hypothesize that this is related to the information capacity of the context versus that of the future. A context with highly informative language will likely take precedence over its respective future when choosing a direction for the story. Techniques to strengthen the future signal or weaken the influence of context could yield better results against a wider array of context and prompt pairings.

\subsection{Summary of Contributions}
In this work we propose a new task within Automated Story Generation called \textit{future conditioning}. Future conditioning is the process of coercing a generative language model towards temporally interpolating between a story context and a future plot event with the goal of integrating it into the story. By providing a generative language model with a story context and a future event, the model is responsible for generating the intermediate details in the form of complete sentences. To this end, new plot events can recurrently be integrated into the story.

In pursuit of our proposed task, we developed \textit{Future Sight}, a model architecture composed of two large pretrained language models connected by a nonlinear hidden layer. By encoding a future plot event with a pretrained transformer encoder and allowing a pretrained transformer decoder to attend to it, we demonstrate that it is possible to dynamically guide pretrained transformers in generating stories.

\bibliography{custom}
\bibliographystyle{acl_natbib}




\end{document}